\documentclass[journal=acscii,manuscript=article]{achemso}

\usepackage[version=3]{mhchem} 
\usepackage[T1]{fontenc}       
\usepackage{url}

\author{Juno Nam}
\author{Jurae Kim}
\email{jurae.kim@snu.ac.kr}
\affiliation{Seoul Science High School, Seoul 110-530, Korea}

\title[]{Linking the Neural Machine Translation
and the Prediction of Organic Chemistry Reactions}

\begin{document}


\begin{abstract}
Finding the main product of a chemical reaction is one of the important problems of organic chemistry. This paper describes a method of applying a neural machine translation model to the prediction of organic chemical reactions. In order to translate `reactants and reagents' to `products', a gated recurrent unit based sequence--to--sequence model and a parser to generate input tokens for model from reaction SMILES strings were built. Training sets are composed of reactions from the patent databases, and reactions manually generated applying the elementary reactions in an organic chemistry textbook of Wade. The trained models were tested by examples and problems in the textbook. The prediction process does not need manual encoding of rules (e.g., SMARTS transformations) to predict products, hence it only needs sufficient training reaction sets to learn new types of reactions.
\end{abstract}

\section{Introduction}

Predicting major products of chemical reactions is a basic issue in organic chemistry. Because the ability to make accurate predictions of products plays a key role in applications such as designing syntheses, enhancing this ability has been one of the major objectives in organic chemistry curricula. The use of computational methods to achieve this ability facilitates highly efficient planning of organic syntheses. There is a strong link between predicting reactions and problem of retrosynthesis as these two are the inverse processes of each other. Therefore various methods predicting reactions with retrosynthesis have been developed during the past few decades. These prediction methods are widely covered in recent reviews in computer-aided organic synthesis planning.\cite{Szymkuc2016, Todd2005}

Current computational methods for predictions of reactions in organic chemistry are generally classified into three categories.\cite{Kayala2011, Kayala2012} The first category predicts the reactions according to rules encoded by humans. Starting from seminal works in this area such as CAMEO\cite{Jorgensen1990} and EROS\cite{Hollering2000} systems, some algorithms based on this method have been developed along the years. For instance, some algorithms identify reactive sites.\cite{Satoh1995, Sello1992} Recently, Chen et al.\cite{Chen2008, Chen2009} presented a prediction system based on the reaction mechanism, using manually composed transformation rules of each mechanistic step. These methods perform well on predicting target reactions included in the composed rules but needs further encoding when new reactions---which are not included in composed rules---are discovered. Because of this need for manual encoding, old projects in this area are already outdated.

The second category is based on physical calculations.\cite{Behn2011, Benko2003, Chaffey-Millar2012, Olsen2004, Plessow2013, Socorro2005, Wang2016, Zimmerman2013} To predict the products, these methods calculate the energies of transition states from plausible reaction pathways. Some methods for choosing reaction coordinates or searching relevant transition states have been developed.\cite{Zimmerman2013} However, because the majority of energy calculations involve quantum mechanics, these methods usually require high computational cost. Various approaches have been developed to mitigate this problem. For instance, ToyChem\cite{Benko2003} calculates energy employing a simplified version of Extended H\"{u}ckel Theory, while ROBIA\cite{Socorro2005} implements rule--based decision process to filter reactive sites.

The last category predicts reactions using machine learning. Although there were several early approaches,\cite{Gelernter1990, Rose1990} this area of reaction prediction has been revisited because of the recent development of machine learning algorithms. For example, ReactionPredictor\cite{Kayala2011, Kayala2012} predicted reactions regarding each step of reaction mechanisms. In this work, the interactions between molecular orbitals were marked with recursive notations to express reaction mechanisms. Neural networks were applied to filter source/sink molecular orbitals and to rank generated pairs of molecular orbitals to yield possible mechanistic steps. On the perspective of general reaction predictions, this kind of method requires tree search to discover the product in account to its scope of mechanistic step prediction. However, in some reactions, each step is not always the most plausible mechanism given by theories, and some reactions include complicated mechanistic pathways. Hence, it could be computationally demanding to predict the overall path of reaction.

Another approach utilizing machine learning is to generate a reaction fingerprint to predict the reaction class by considering the whole reaction, not the detailed mechanism. Schneider et al.\cite{Schneider2015} developed a fingerprinting method for classification of reactions based on molecular fingerprints such as AtomPairs,\cite{Carhart1985} Morgan\cite{Morgan1965}(ECFP\cite{Rogers2010}), and TopologicalTorsions.\cite{Nilakantan1987} Reactions were classified according to reaction types included in the RSC's RXNO ontology.\cite{RXNO} Focusing on the classification, this work had used products to predict the reaction types. Recently, Wei et al.\cite{Wei2016} developed a reaction type classifier that is dependent only on reactants and reagents, not products. They also applied the differentiable neural fingerprinting method\cite{Duvenaud2015} to generate reaction fingerprints, and combined the classifier with SMARTS\cite{JamesC.A.} transformations to yield the product molecules. This process of predicting products by reaction classes and corresponding SMARTS enables predicting products which match to those classes to be relatively easy. On the other hand, this method needs manual encoding of SMARTS transformation for new kinds of reactions. In addition, classification of reactions sometimes suffers from ambiguous reaction classes. Due to these drawbacks, an alternative approach to predict reactions without classification is needed.

\begin{figure}
\begin{center}
\includegraphics[width=\columnwidth]{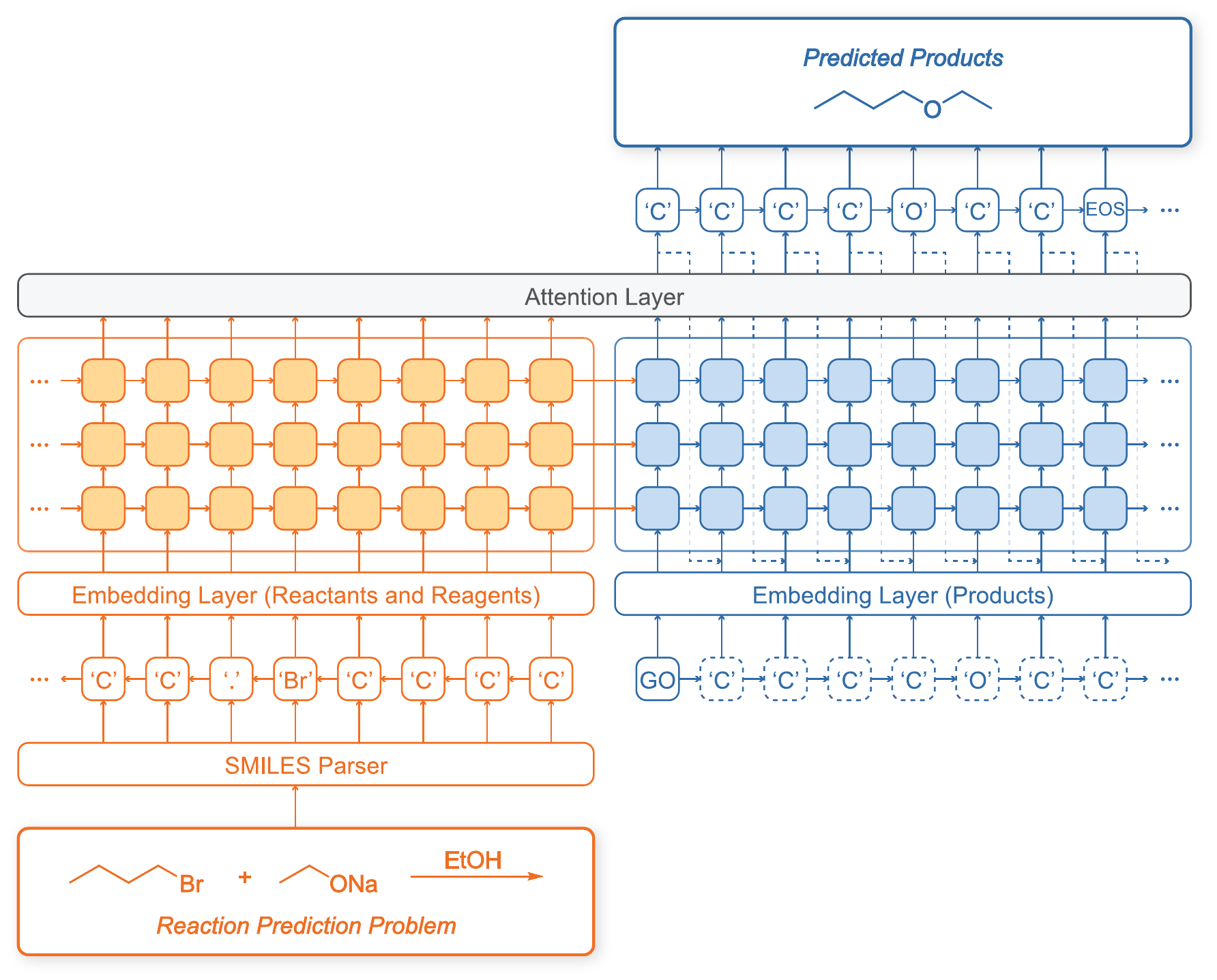}
\caption{\label{fig:1} An overview of this paper's method for product prediction. Reactants and reagents are converted to SMILES strings, tokenized by SMILES parser, and reversed. Each token is transformed into an embedding vector, and provided as an input to the encoder--decoder sequence--to--sequence model with attention mechanism, which is comprised of three GRU\cite{Cho2014} (Gated Recurrent Unit) layers. Generated tokens are concatenated to build up a product prediction.}
\end{center}
\end{figure}

Although reactant, reagent and product molecules involved in reactions are three-dimensional entities, they can be represented as linear SMILES\cite{Weininger1988} strings, which can be decomposed to a list of atoms, bonds and several kinds of symbols. Hence, in a linguistic perspective, SMILES can be regarded as a language with grammatical specifications. In this sense, the problem of predicting products can be regarded as a problem of translating `reactants and reagents' to `products'. In this work, a sequence--to--sequence translation model is applied to predict reactions. A parser to tokenize the SMILES strings was constructed, and two kinds of training and test sets were generated: one reaction set from manually encoded rules, and another set from real reaction database. An encoder-decoder model of neural translation was trained using these reaction sets to predict the correct products. An outline of the prediction system is shown in Figure \ref{fig:1}.

\section{Results and Discussion}

After applying the training set generation process explained in the Methods section, two training sets were generated: one from the patent database, and another one from the reaction templates in an organic chemistry textbook of Wade\cite{Wade2013}. Each training sets will be subsequently mentioned as `real' and `gen' training set. Utilizing those training reaction sets, two reaction prediction models were built: one model using the `gen' training set, and another one using both the `gen' and the `real' training sets. Those two models are compared to investigate the effect of the `real' training set on the reaction prediction model.

\subsection{Performance on textbook questions}

\begin{figure}
\begin{center}
\includegraphics[width=0.8\columnwidth]{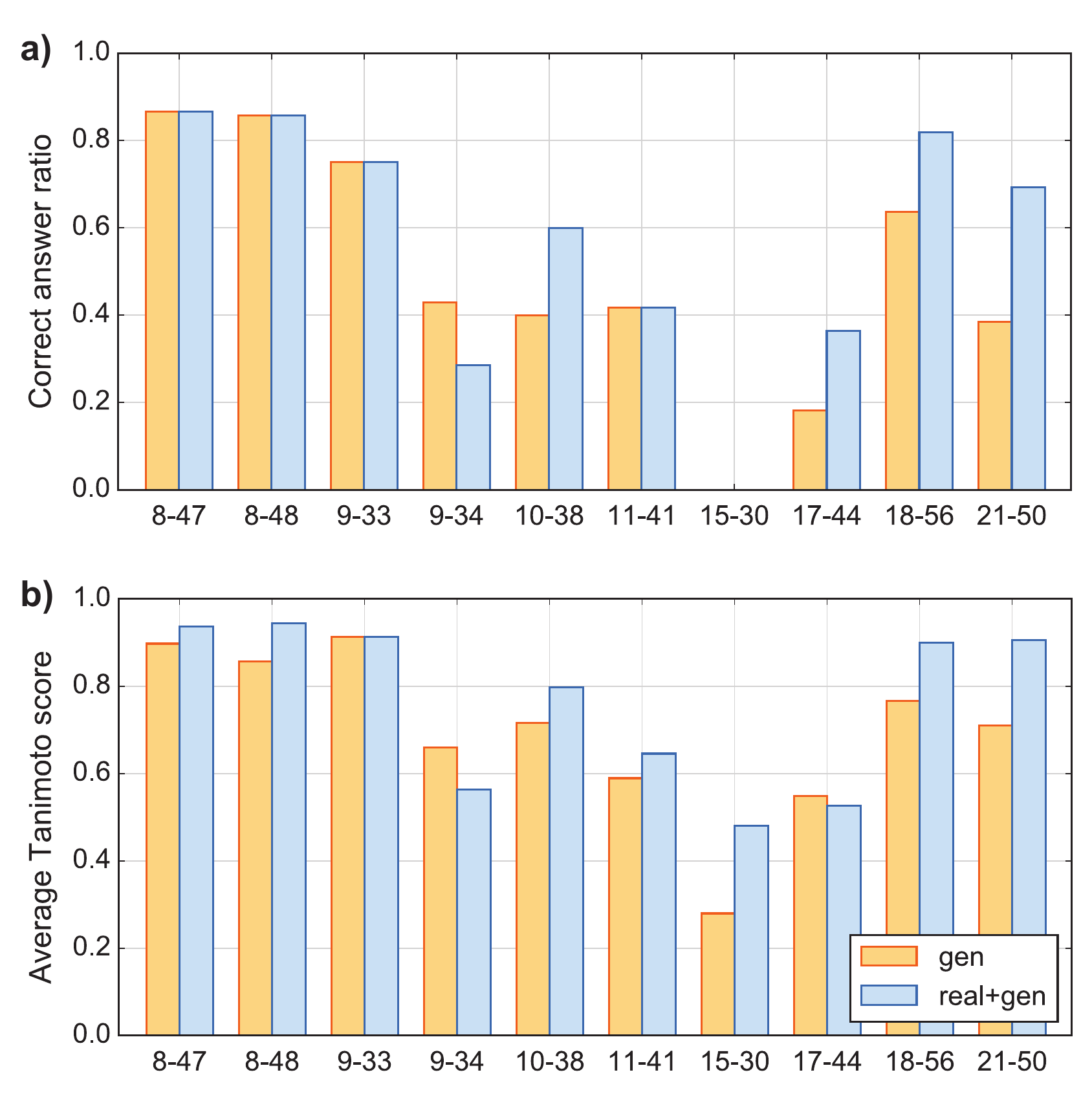}
\caption{\label{fig:2} Prediction results for organic chemistry problems in Wade.\cite{Wade2013} (a) Ratio of fully correct predictions (b) average Tanimoto score in each problem set.}
\end{center}
\end{figure}

To test the trained models, problems in Wade\cite{Wade2013} were applied, following the method of Wei et al.\cite{Wei2016} 10 problem sets from the textbook were applied. Every problem is treated as a product prediction problem, and problems out of scope of this work, such as simple deprotonation, were excluded from the problem set. Each problem set is constituted of 6 to 15 reactions. For every problem in each set, the problem reaction is converted into the reaction SMILES string, and the product part is removed. This product--less SMILES string is fed as an input to the two reaction model, and models (`gen' and `real+gen' model) produce the product SMILES strings. This produced product is compared with the original product to evaluate each model. The ratio of correct answers and the average Tanimoto\cite{Bajusz2015} similarity between Morgan fingerprints of the predicted products and Morgan fingerprints of the real products were used as evaluation metrics. The product generation is based on the tokenized SMILES string symbols, so this process can sometimes generate invalid SMILES strings, such as not closing the opened branches (mismatched parentheses). If generated product SMILES string contains such errors, the score for corresponding prediction was set to 0. The overall prediction results are shown in Figure \ref{fig:2}.

Comparing two models, data in Figure \ref{fig:2} shows that the prediction ability of the `real+gen' model is better than the `gen' model in most cases. It is evident from the results that the training set from the real patent reactions facilitates the product prediction procedure. Training set from the generated reactions does not include reactants with either more than 10 atoms or multiple functional groups. However, the test problem sets include such reactants, and the reasonably good performance on these test problems indicates the generalizability of this model. Problem set 15-30 regards Diels--Alder reactions and 17-44 regards the reactions with benzene as the reactant, hence the reactions in these problem sets are not in the scope of training sets of the generated reactions. For problem set 15-30, though both models didn't predict the fully correct answers for all 6 problems, the `real+gen' model retrieved better results on the average Tanimoto score. The low correct answer ratio on Diels--Alder reactions could be due to the lack of simple training data for those reactions. Although Diels--Alder reactions are included in the training set from the patent data, they are rather complex. Hence features of Diels--Alder type reactions may be suppressed when training the model regarding these sets of reactions. The `real+gen' model's better result on the Tanimoto score could account for the lower ratio of invalid product SMILES strings, because the `real+gen' model was trained on larger number of reactions than the `gen' model. Larger number of training sets may have resulted in decoder networks generating more valid SMILES strings. For problem set 17-44, the `gen' model correctly answered two, while the `real+gen' model correctly answered four out of eleven test problems. Reactions of aromatic compounds are only included in the `real' training set, thus it is reasonable that the `real+gen' model yielded relatively better prediction results. However, the `gen' model correctly predicted two reactions, implying that this prediction model even can extrapolate into the unencoded reaction patterns.

\subsection{Scalability of the model}

\begin{figure}
\begin{center}
\includegraphics[width=0.8\columnwidth]{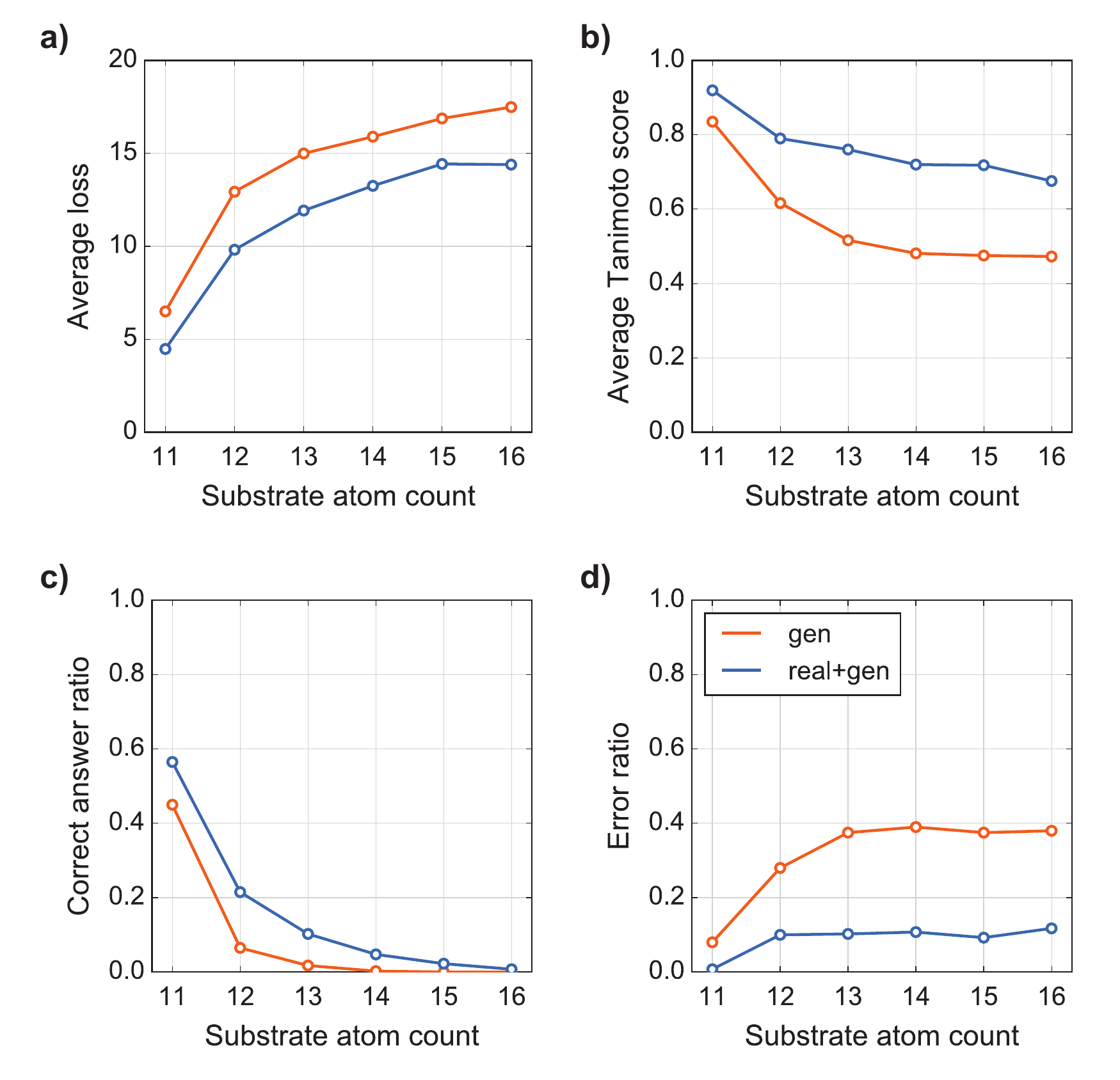}
\caption{\label{fig:3} Prediction results for the generated test reactions according to atom count of the substrates used in reaction generation. (a) Average cross--entropy loss in the models (b) average Tanimoto score (c) ratio of fully correct predictions (d) ratio of invalid product SMILES strings generated in each test reaction set with corresponding substrate atom count and model.}
\end{center}
\end{figure}

To further test the models of their scalability, test reactions are generated with a similar method from generating the training set of manually composed reactions. Substrates used when generating test reactions are molecules with 11 to 16 atoms with a single functional group, while the training set includes molecules with under 11 atoms. Thus, the GDB-17\cite{Ruddigkeit2012} molecule dataset was used instead of GDB-11 to generate substrates. The generated test set consists of a total of 2400 reactions, 400 for each substrate molecule atom count of 11 to 16. Evaluation metrics include ratio of correct answers, average Tanimoto score calculated in the same way as the test set of textbook problems, average cross--entropy losses in the models, and the ratio of errors (invalid SMILES string). The prediction results are shown in Figure \ref{fig:3}.

As illustrated by the Figures \ref{fig:3}b and \ref{fig:3}d, the `real+gen' model generally performs better than `gen' model when predicting from longer sequences of reactants and reagents. The `real+gen' model is considerably lower in their error ratio of invalid SMILES string generation compared to the `gen' model. The `real + gen' model maintains a Tanimoto score around 0.7 and an error ratio around 0.4, when the number of atoms in substrate molecules increases. Despite this, the ratio of the fully correct prediction decreases quickly as the number of atoms in the substrate molecules increases. As the SMILES string lengths of the reactants and products are proportional to the atom count of the substrates, this result reveals that the recurrent network models in this work generate mistakes on prediction when the input sequence lengths become longer.

\begin{figure}
\begin{center}
\includegraphics[width=0.6\columnwidth]{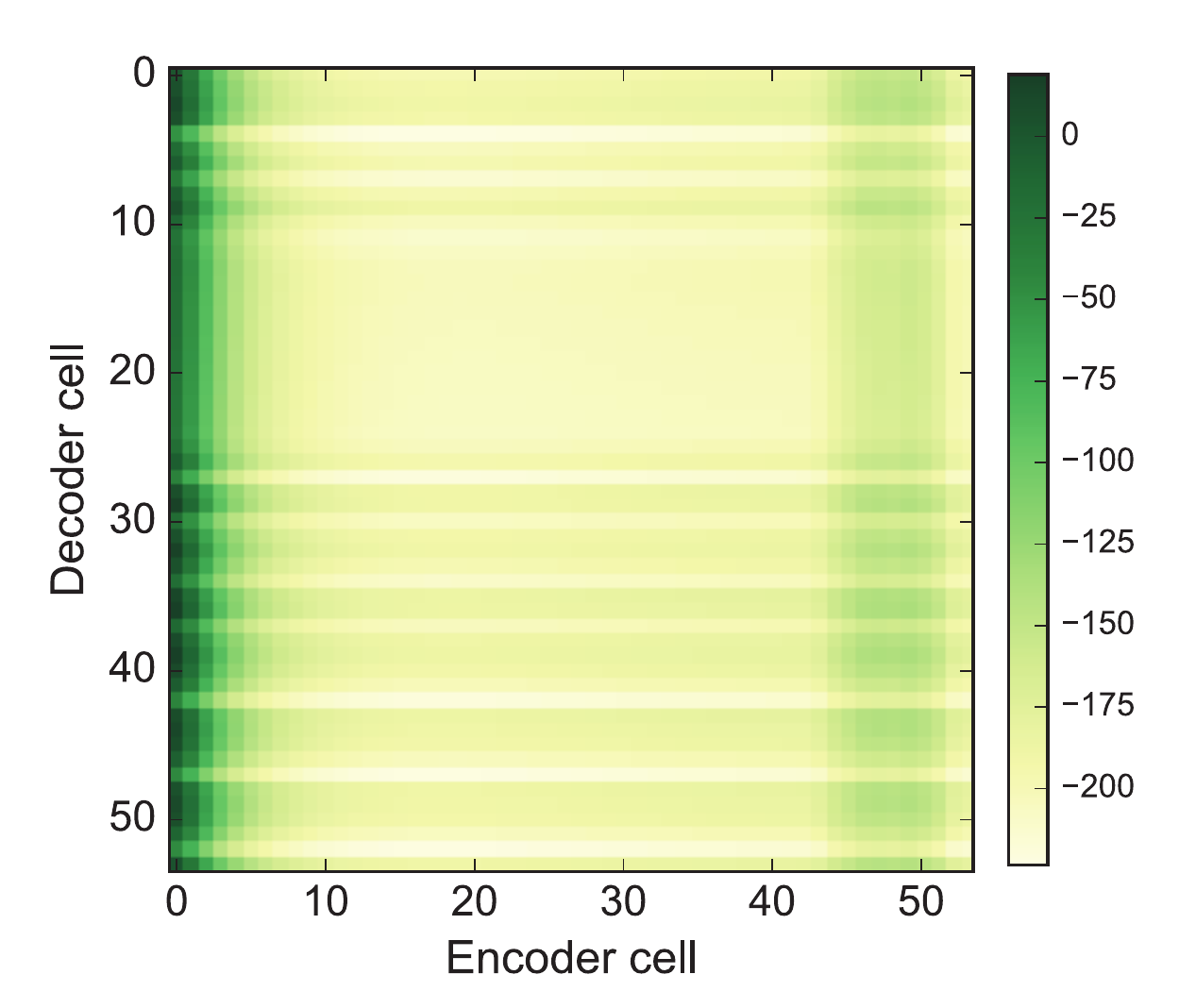}
\caption{\label{fig:4} The vectors (in each decoder cell) which generate the attention weights when the softmax function is applied. The vectors were extracted from the `real+gen' model while generating the predicted product SMILES string `CCC(C)(C)Cl' after given the reaction problem  `CC=C(C)C.Cl\textgreater{}\textgreater'.}
\end{center}
\end{figure}

\begin{figure}
\begin{center}
\includegraphics[width=0.9\columnwidth]{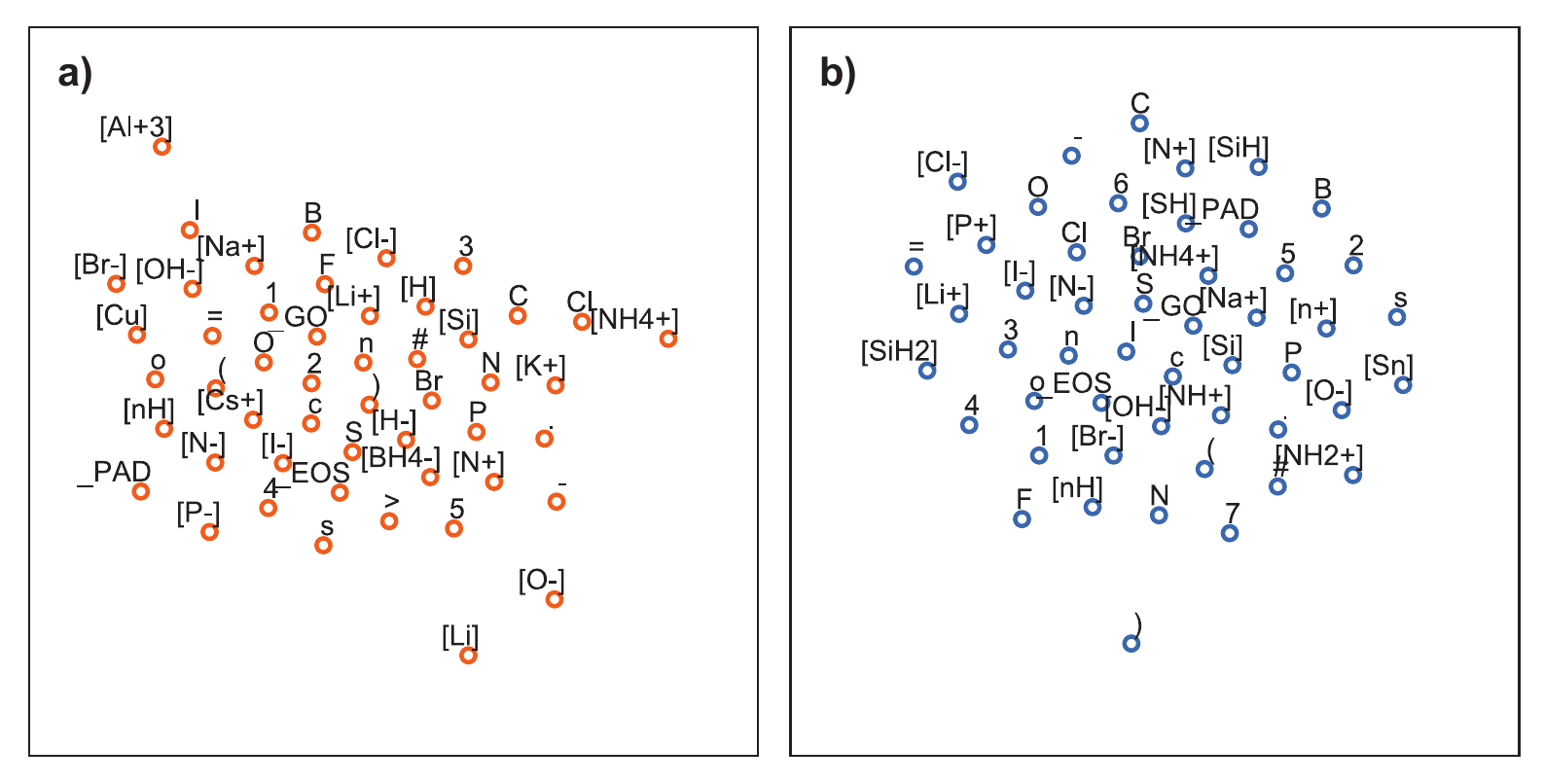}
\caption{\label{fig:5} The embedding vectors of (a) encoder and (b) decoder tokens, visualized on two dimensional space by t-SNE\cite{VanDerMaaten2008} (t-Distributed Stochastic Neighbor Embedding) algorithm with perplexity of 50. The most common 50 tokens in each token set are showed.}
\end{center}
\end{figure}

The vectors (in the decoder cells) which generate the attention weights are visualized in Figure \ref{fig:4}. As illustrated in the figure, values corresponding to the first few encoder cells are significantly higher than other values in the vectors. Hence, when the softmax function is applied to generate the attention weights, the first elements of the weight vectors become close to 1, and the rest to 0. This results in the decoder cells only attending to the first encoder cells. Attention mechanism is generally adopted in neural translation models to enhance `alignment', which means decoder attending to related encoder cells. If the decoder cells which generate the tokens representing the unreactive sites in the reactant molecules can attend to the correlating encoder cells, analogous to the atom mapping, the reaction prediction with longer input sequences or larger numbers of reactant atoms can be improved. Additionally, as this work is focused on training and testing a translation model for predicting the main product(s) of the reaction, the chemical natures of the participating molecules are overlooked. For instance, the embedding vectors of each token in the encoder (reactants and reagents) and decoder (product) are not related to their chemical features, as visualized in Figure \ref{fig:5}. Also, the scope of `main product(s)' can be ambiguous, such as regarding or not regarding the leaving protecting groups from the deprotection reactions. Future works on this type of reaction product prediction algorithm will have to address this issue for improvements.

\section{Conclusion}

This work have dealt with the application of neural machine translation in the field of organic chemistry reaction prediction. Two models (the `gen' model and the `real+gen' model) were composed, and the comparison of results between two models showed that the training on real reaction facilitates the prediction ability of the model. The models predicted the products of the reactions in a reasonably high precision, and in the case of the `gen' model, the model could extrapolate their prediction ability to untrained types of reactions (reactions with aromatic substrates). While the test sets used to acquire quantitative results were elementary reactions, the `real+gen' model was able to predict some high--level reactions because it was trained on the recent patent reaction.

Comparing with previous works applying machine learning to reaction prediction task, the mechanism--based model of Kayala et al.\cite{Kayala2012} is better on reactions with single mechanistic step, while only few multistep reactions were shown on their work, as those reactions need tree--search algorithm to discover the mechanistic pathways to final products. This work used similar training set generation and evaluation metrics with the fingerprint--based model of Wei et al.,\cite{Wei2016} and the model in this work performed better on product generation in test set of organic chemistry textbook questions. Also, this algorithm generates product SMILES strings from tokens; hence manual input of SMARTS transformations is not needed. This allows overall process to be significantly flexible, as this method only requires sufficient data of reactions to train on. However, this also generates some problems such as composing invalid product SMILES strings, and reactions with multiple pathways---for instance, substitution and elimination---are hard to deal with present model structure. Future version of this algorithm should deal with these problems.

Machine learning based reaction prediction algorithms are able to generate the prediction much faster than the methods using quantum calculations, making these algorithms suitable for synthesis planning tasks.\cite{Szymkuc2016, Wei2016} The reaction prediction models can be extended to deal with the quantity of reagents, reaction temperature and time to gain more elaborate prediction result, or it can be used to predict yield of the possible products. The link between this method of reaction prediction and machine--assisted chemistry\cite{Ley2015} will be able to open the new areas of automatic chemical systems.

\section{Methods}

\subsection{Training reaction set composition}

Two training sets were generated to train the reaction predictor model. The first set is based on real reactions. There exists reaction databases such as CASREACT\cite{Blake1990}, Reaxys\cite{Reaxys}, or SPRESI\cite{Roth2005}, but they are commercial databases, and the reactions included in these databases cannot be extracted as appropriate forms for this work. Hence, the reaction database collected from patents by Lowe\cite{Lowe, Lowe2012} was used. Schneider et al.\cite{Schneider2015} had also used this database to train reaction classification system. In this work, reactions extracted from 2001--2013 USPTO applications were used. First, atom mappings were removed from the reaction SMILES as they are unnecessary for the translation model. To filter out inappropriate reactions for the translation model, (1) reactions with reactants and reagents lengths (length of string before the second `\textgreater' in reaction SMILES) longer than 150, (2) reactions with products which lengths (length of string after the second `\textgreater') are longer than 80, and (3) reactions with four or more products were excluded. A total of 1,094,235 reactions were collected.

Because the reactions are relatively new, this reaction set lacks elementary reactions. Hence, following the method of Wei et al.,\cite{Wei2016} the second reaction set was composed according to elementary reactions in an undergraduate organic chemistry textbook by Wade\cite{Wade2013}. A total of 75 reaction types regarding five types of substrate molecules (acid derivatives, alcohols, aldehydes and ketones, alkenes, alkynes) were considered. For each reaction type, reactions were generated by iterating the reactant molecules which match the reaction template specified as a SMARTS transformation. Reactant molecules with 1--10 atoms were extracted from the molecule database GDB-11.\cite{Fink2005, Fink2007} As all halides in GDB-11 are fluorides, F was substituted to Cl, Br, I in each halide to generate alkyl halide reactants. Molecules with either multiple functional groups or bulky groups such as neopentyl group were excluded. RDKit\cite{RDKit} was used to collect matching reactant molecules and generate reactions from the reaction template. A total of 865,118 reactions were generated in this way.

\subsection{Reaction predicting translation model}

Recently, encoder-decoder models are widely used to solve sequence--to--sequence tasks such as translating sentences.\cite{Bahdanau2014, Jean2015, Johnson2016, Luong2014, Sutskever2014, Vinyals2014, Wu2016} Recurrent neural networks like LSTM\cite{Hochreiter1997} (Long Short-Term Memory) or GRU\cite{Cho2014} (Gated Recurrent Unit) are used to deal with inputs and outputs in arbitrary length. From a mathematical perspective, translation is to find the result sentence $\mathbf{y}$ which maximizes the conditional probability $p(\mathbf{y}|\mathbf{x})$ when given the input sentence $\mathbf{x}$. This probability is the product of conditional probabilities for each token in sentence $\mathbf{y}$. Hence, the sequence--to--sequence models are designed to generate result sequences by estimating those conditional probabilities.\cite{Sutskever2014} To apply this language model into the reaction prediction problem, chemical reactions have to be represented as lists of tokens. In this work, SMILES strings are used to represent reactions and those strings are tokenized into a list of atoms, branching symbols (parentheses), ring closure numbers, and bonds. Similar method of tokenizing SMILES strings for recurrent neural networks have been used in chemical autoencoder\cite{Gomez-Bombarelli2016} and activity prediction task.\cite{Jastrzebski2016} For simplicity of implementation, a PEG\cite{Ford2004} (Parsing Expression Grammar) based parser was composed to tokenize the SMILES strings using the pyPEG\cite{PyPEG} module in Python, following the method of Smidge\cite{Smidge} (lightweight OpenSMILES\cite{CraigA.James} parser), which is implemented in JavaScript.
Using the tokens parsed from the training set, the numbers of different tokens that have to be used as inputs to the encoder (reactants and reagents) and decoder (products) are counted. According to this, the input (reactants and reagents) and output (products) vocabulary size were decided as 311 and 180. The sequence--to--sequence model used in this work uses 3 layers of GRU cells, and embedment of tokens into vectors of size 600. Model with buckets\cite{Seq2seq} was used to process various lengths of the encoder and decoder input, which means that reactions with input length within the range determined by bucket are processed by the model with corresponding encoder and decoder lengths specified by that bucket. Buckets with (54, 54), (70, 60), (90, 65), (150, 80) ((encoder length, decoder length)) are used in this work. Additionally, attention mechanism\cite{Bahdanau2014} is applied to allow the model to search relevant information in the encoder in every decoding step. The sequence--to--sequence model\cite{Seq2seq} given in the TensorFlow\cite{Abadi} library is applied at this work. Therefore, the rest of the specifications for reaction prediction model in this work, such as the learning rate, follow the default conditions in the TensorFlow library.
To provide consistency of the input and output sequences, each reaction SMILES string fed into the prediction model is first normalized by canonicalizing each substring of reaction SMILES string split by the character `>'. RDKit was used for this canonicalization process. Then, the normalized string is tokenized into the vectors of input tokens (reactants and reagents) and output tokens (products). Input tokens are reversed, embedded into vectors, and provided as inputs to the three layers of encoder GRU cells with length specified by the corresponding bucket. Tokenized SMILES strings of products are used as the decoder inputs when training the model, and the output of the previous decoding step is used when testing the model. Generated decoder output tokens are concatenated to generate the product SMILES prediction.

\begin{acknowledgement}

This work was in part supported by the Korea Foundation for the Advancement of Science \& Creativity. The authors acknowledge Jennifer N. Wei for allowing the use of her codes with useful discussions, Minho Kim for extensive support on the logical structure and English of this manuscript, and Hanjun Lee for useful discussions.

\end{acknowledgement}




\bibliography{bibl}

\end{document}